# Artificial Intelligence and radiation protection.
# A game changer or an update?

S. Andresz[1*], A Zéphir[2], J. Bez[3], M. Karst[4], J. Danieli[5]

[1] Nuclear Protection Evaluation Centre, 92260 Fontenay aux Roses, France
[2] Prevision.io, 75008 Paris, France
[3] IRSN, 92263 Fontenay aux Roses, France
[4] EDF-UNIE, 93282 Saint-Denis, France
[5] SPRA, 92141 Clamart, France
* Corresponding author: sylvain.andresz@cepn.asso.fr

**Abstract.** – Artificial Intelligence (AI) is regarded as one of the most disruptive technology of the century and with countless applications. What does it mean for radiation protection? This article describes the fundamentals of machine learning (ML) based methods and presents the inaugural applications in different fields of radiation protection. It is foreseen that the usage of AI will increase in radiation protection. Consequently, this article explores some of the benefits and also the potential barriers and questions, including ethical ones, that can come out. The article proposes that collaboration between radiation protection professionals and data scientist experts can accelerate and guide the development of the algorithms for effective scientific and technological outcomes.

**Résumé.** – L'Intelligence Artificielle (IA) est une des technologies les plus révolutionnaires de ce début de siècle, génératrice d'applications illimitées. Quelles conséquences pour la radioprotection ? Après un aperçu des techniques actuelles d'apprentissage automatique, cet article présente leurs premières applications dans différents domaines de la radioprotection. Il est probable que ces techniques d'IA seront de plus en plus utilisées. Cet article explore les bénéfices de l'IA, mais aussi les freins potentiels et les questions, notamment éthiques, qui pourront surgir. L'article suggère que l'association entre les professionnels de la radioprotection et les experts de l'IA permettra de tirer les meilleurs fruits des algorithmes.

## 1. INTRODUCTION

The first contact of the Y generation (*i.e.* 30-40 years old) to Artificial Intelligence (AI) was usually trough movies and comics where AI was depicted as a breakthrough invention that can perform tasks far better than human (like the cubicle robot in *Interstellar*), take in tedious works (*Wall-E* cleans garbage), compete in poetry contest (Zéphir, 2019) or empathize with us using impressive conversational skills (Samantha, the OS in *Her*). But more frequently AI transcend mankind and wipe us off from Earth (Skynet in the *Terminator* series) or at least put us in slavery (*The Matrix*). The older generations certainly recall *2001: A Space Odyssey* where computer HAL is taking control of the ship or the murderous human-replicas in *Blade Runner*.

Although movies and comics tend to exaggerate for dramatic purposes, there might be some truth behind. After impressive achievements, AI, Machine Learning, Neural Network and Big Data have become popular terms in media and are promised to be everywhere in the future, the nuclear sector being no exception (Gomez-Fernandez, 2020, NEA, 2021). Therefore, knowing what AI is actually and its capacities seems crucial.

In the framework of a global reflexion about innovation and the future of radiation protection (RP) ignited by (Bourguignon, 2017) and continued in (Ménard, 2019), the Young

Club of the French Society for Radiation Protection has decided to assemble a working group "AI & RP" (in March 2020) to further investigate this topic.

To obtain a picture of the landscape, the working group has performed a bibliometric analysis of publications on "artificial intelligence" and synthetized publications about the usage of AI in RP and its allied fields. Unformal interviews with RP professionals engaged in AI projects and Data Scientists were also conducted. The purpose of this article is to present the result of this working group.

## 2. WHAT IS ARTIFICIAL INTELLIGENCE?

### 2.1 And what is Machine Learning?

The first issue was the meaning of "artificial intelligence". The term was introduced in 1956 (Minsky, 1963), so it is almost as old as the computer itself, but its definition has always been loose, debatable and evolving with the techniques. Two branches of AI have co-existed: symbolic (or logic) IA which import factual and heuristic knowledge of human experts to achieve the solution was popular in the 50's and the 80's while numerical AI has followed a different path: rather than explaining to the computer how to solve a problem, Machine Learning (ML) "*gives computers the ability to learn from and improve with experience, without being explicitly programmed*" (Parliament, 2018).

One feature of ML is the 'artificial neuron', an original informatic architecture mimicking the functioning of the brain. The simplest artificial neuron receives inputs from several 'synapses', weight them, calculate a score and passes an output if above a threshold/activation function. The trick is that some parameters of this algorithm are adjusted during a learning phase by the backpropagation of the distance between the output and the 'true' result. A Neural Network (NN) is constituted with more than 3 layers of neurons, inclusive of the inputs and outputs. Training multiple layers of neurons constituted the basic of Deep Learning (DL).

The artificial neuron was theorized precociously in 1943 and built in 1957 (Rosenblatt, 1958) but this branch of IA suffered from technical limitations and faced a lack of innovation and investment for almost 30 years. Particularly in France, this technique faced reluctance for not producing enough predictable outputs and not being 'cartesian' enough (ENS, 2018).

The U-turn came around 2012: the availability of massive banks of data (Big Data) for training the NN and the innovations in graphical process units (GPU), initially designed to make numerous simple calculations in parallel, have unlocked the chains of DL. Further advancements include the number of layers (up to hundreds now) or Convolutional NN (NN including pooling and convolution functions). In the media, DL probably pinnacle when AlphaGo defeated a human Go player in 2016. The mediatic buzz has weaken, but soon, policy makers have turned their attention to the topic and laid down strategies for ML and DL (often generally referred as AI) (Parliament, 2018), France being no exception (Villiani, 2018).

We decided to focus on ML techniques; one of the most popular achievement of this technique today is image recognition.

### 2.2 Fundamentals of Machine Learning

Simply put, ML is an ensemble of algorithms producing a model from data, the key feature being the algorithm learns the model from the data and do not use pre-programmed rules. Implicitly, the model should be difficult to describe: complex, non-linear phenomena, multiple inter-related parameters (or dimensions). The inferred model is typically used for predictive analysis, ML can be also used to approximate function or explore (large) data set. All types of data can fit: time series, image, text, measured or computed etc.
- Supervised learning is one form of ML, using data labelled by humans, for classification purposes (assign a category) or regression (to predict a numerical value). Popular algorithms include linear classifiers, support vector machines (SVM) or decision tree. On the downside, the models can be time-consuming to train, and the labelling require human time and expertise.
- Unsupervised learning lets the algorithm alone in charge to find structures: aggregate of resembling data (clustering) making use of K-Means or DBSCAN algorithms or trends/rules in data (association) ex. with naïve Bayes. Unsupervised learning models are often computationally complex as they need a large training set to produce intended outcomes.
- In reinforced learning, the algorithm adjusts its weight based on its observation of the impact of its output on the environment to make a different decision next time. AlphaGo was trained by playing Go against other programs.

Not limited to, the preparation of the data includes: collection, harmonization, outlier's exclusion, normalisation. The data must be separated between the training, validation and testing sets, in a manner to avoid bias, underfitting (the algorithm does not converge) and overfitting (the algorithm cannot generalize to data outside the learning set).

Different data preparation, algorithms and metrics (cost function, performance etc.) will be tested to design the most suitable ML algorithm for the problem under consideration. The expression "*no free lunch*" used by Data Scientists expresses the fact that no algorithm can solve all the issues: if one algorithm works well against one problem, it will need adjustments to work against another, even comparable. Finally, the prototype will be tested (eventually fixed) and deployed. Altogether, a ML project is like any engineering task (Prevision.io, 2020). The Data Scientist cannot work alone: most steps require discussion and collaboration with the experts in the field

## 3. BIBLIOMETRIC ANALYSIS

We performed a bibliometric analysis to evaluate the number of yearly publications on the applications of ML in various scientific fields and then focused on the application of ML in the "radiological" field. We used Scopus for its larger coverage and accuracy (Falagas, 2008).

Scopus was accessed in June 2021 with the following research strategy: TITLE_ABS_KEY = ("Machine AND Learning AND *Field*"), PUB_YEARS > 2010; with a different '*Field*' entered for each research. The method identifies publications with titles, abstracts or keyword with the terms in brackets published after 2010. There is a delay between the acquisition of results and their publication and some researches are not published in English, but screening the 2011-2020 period should be informative. The results for 7 '*Fields'* are presented in Figure 1.

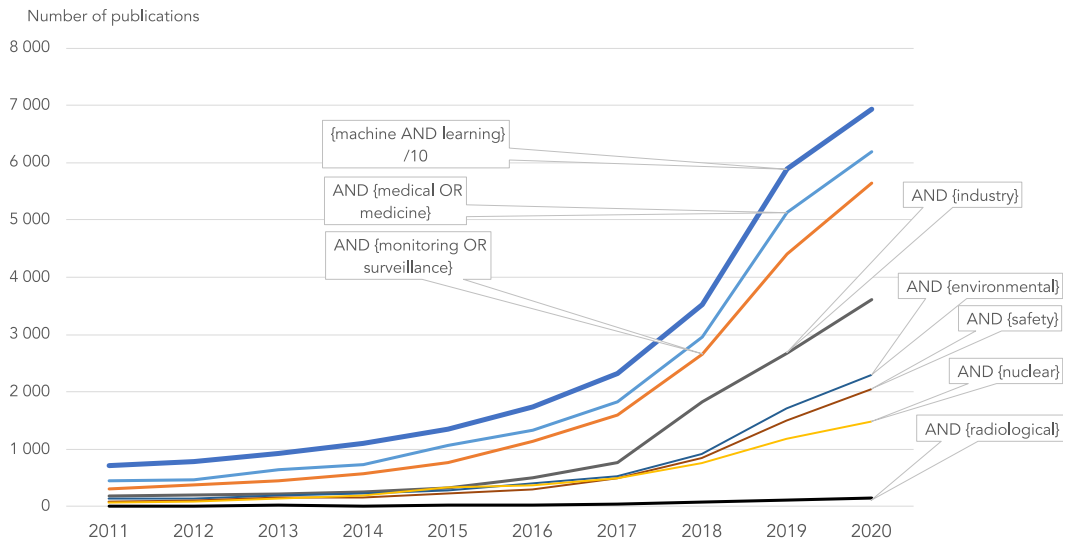

Figure 1. Number of publications/year on applications of Machine Learning in various fields.

The usages of ML in the "medical/medicine" sector and for "monitoring/surveillance" meets the eyes by the number and rate of publications, with notable accelerations after 2016 and then after 2018. ML seems to have found applications in the "industry" field after 2017, but less numerous and at lower pace. Then comes a group "environmental", "safety" and "nuclear" fields where the interest has started after 2018 and the total numbers and trends in publications are more modest than the above-mentioned fields. It seems that the interest has hardly started in the "radiological"[1] field.

When focusing on "radiological', the subject areas (SUBJ_AREA), the country (AFFIL_COUNTRY) and the techniques (identified in the KEY_WORD) of the N=469 results have been collected and are presented in Table 1.

**Table 1.** Number of publications on applications of Machine Learning in radiation protection by subject area and by country and techniques.

| TITLE_ABS_KEY(Machine AND Learning AND radiological), PUBYEAR > 2010 | | | | | |
|---|---|---|---|---|---|
| SUBJ_AREA | Number of publications | AFFIL_COUNTRY | Number of publications | KEY_WORD | Number of publication |
| Medicine | 333 | USA | 176 | ML | 118 |
| Computer Science | 127 | China | 67 | DL | 112 |
| Biochemistry, genetics | 100 | Germany | 43 | Convolutional NN | 53 |
| Engineering | 70 | United Kingdom | 41 | Classification | 46 |
| Health Professionals | 45 | India | 30 | Artificial NN | 38 |
| Neuroscience | 36 | Italy | 28 | NN | 41 |
| Physics and Astronomy | 32 | Canada | 27 | SVM | 30 |
| Mathematics | 20 | France | 23 | - | - |
| Multidisciplinary | 16 | Switzerland | 18 | | |
| Material Science | 15 | Spain | 17 | | |

---

[1] "*Radiation*" produced out-of-the-scope results (ex. researches on thermal, astronomy, energy etc.) and "*radiation protection*" yielded not enough result (< 60).

The results show the 'hot areas' where ML and "radiological" have met. The prevalence of the medical sector is undebatable (65% of the total of publications with Medicine, Biochemistry, Health Professionals and Neuroscience), followed by a group in fundamental researches (21 % with Computer Science, Mathematics, Physics and Astronomy) and then applied sciences (13% with Engineering, Material Sciences and Multidisciplinary). The USA are in the forefront in publishing, followed at far distance by China, and then a peloton composed mainly by European countries and France ranked at $8^{th}$ place. When it comes to the techniques deployed, DL is the most prominent (although this result is based on the limited perimeter of the keywords attached to each article).

We then analysed several publications where ML has been used for radiation protection purposes, giving priority if possible to achievements in the French context. The synthesis of the analysis is presented in part 4 and a discussion is proposed in part 5.

## 4. MACHINE LEARNING AND RADIATION PROTECTION: EXAMPLES OF APPLICATION

This part presents examples of applications of ML in radiation protection and its allied fields (without pretending to be exhaustive). The initial aim was to focus on French publications (published by French teams or institutions) however we extended this scope for some fields due to the scarcity of the initial results.

### 4.1 The medical sector

#### 4.1.1 Medial image recognition

ML has become famous for pattern recognition, an apropos capacity for medical image interpretation. Examples of concrete applications reported in France are listed by Malchair, 2020: detection of abnormal/hard-to-detect findings, classification of nodule or tumours. AI (in this case: ML) for cancer detection was one topic of the last INCA congress (JSD, 2020) and the focal topic of the annual seminar of the Radiologists Society (JMR, 2018). The press regularly echoes usages of AI in the field (AP-HP, 2019).

The French Association of Medical Physics has embedded AI in its last annual congress (SFMP, 2021). At European level, the Federation of Medical Physics has even published a focus issue (EFOMP, 2021) where several meta-analyses corroborated that image recognition is the prevalent usage of ML in RP (70 % of the publications in the Italian context) and that all imaging practices using radiation are concerned, with a particular attention on lung cancer diagnosis (not to mention the use of AI to screen the lung in CT of Covid patients (Glangetas, 2020)). In a second row, AI-based techniques have been used for image quality improvements (denoising), segmentation and the reconstruction of 2D/3D images and it has been considered to use AI to elaborate personalized low-dose imaging protocols rather than using generic sets (Lewis, 2019).

By (pre)processing the huge number of images generated in the workflow of clinical practice, ML has optimized time for the staff to concentrate on the problematic cases. By improving image quality and/or interpreting image from low-doses ones, ML could have assisted staff in reducing the exposure of patients (EFOMP 2021).

#### 4.1.2 Definition of treatment

ML techniques have been used for facilitating radiotherapy by contouring the organs to irradiate (Hu, 2019) or by predicting the dose distribution in the tissue (Barragan Montero, 2021), therefore reducing exposure to patients by "*increased precision and efficiency*"

(Deutsch, 2021). A specific topic is ML prediction of normal tissue injuries from radiation which constitute a topic of a congress French Radiation Oncology Society (SFRO, 2020).

The prediction patients' response to immunological treatment of cancer (rather than using radiation) has also shown success at Institut Gustave Roussy (*ibid.*). It is also proposed to train AI to 'read' medical and non-medical data and predict the occurrence of cancer (CHU-BDX, 2019).

### 4.1.3 Radiobiology and epidemiology

Radiation biology aims to quantify the risk after an exposure, although the number of biological manifestations, their interactions and co-founding factors make the response difficult to grasp. Initiatives to use ML to capture the dose-effect relationship have been made at different scales: single-cell (Milliat, 2020), tissues and organs (Indraganti, 2019).

In epidemiology, ML (naïve Bayes) has been developed to estimate the sanitary impact of co-exposures on cohort of workers and the authors have stressed the "*novelty and richness of the results*" (Ancelet, 2021).

## 4.2 Metrology

### 4.2.1 Radionuclides identification and detection of event

Isotopes identification in the energy peaks of spectroscopy has been a hotbed for ML. Gomez, 2021, building on previous literature, presented a general methodology for spectral un-mixing with NN. A notable feature of this work is to incorporate explainability method (saliency map) to visualize where/how the algorithm made its decision. In France, Xu *et al.* have carried on similar researches showing promising results for timely identification of radionuclides (Xu, 2020), reduction of uncertainties and limitation of false-positive (Xu, 2021).

It is proposed to incorporate these proofs of concept in surveillance systems such as atmospheric monitoring (ex. the French OPERA-AIR; IRSN Press, 2021a), gamma dose rate monitoring, (ex. French Teleray, Chojnacki, 2021) and radiation monitoring portal at harbour (Koo, 2020). ML is also intended to be used by military for the triage and management of victims in the case of radiological event (SSA, 2021).

### 4.2.2 Modelling

Several attempts to use ML in the fields of instrumentation and environmental modelling can be reported.

In France, Darley has used NN to learn the calibration function of public-available detectors to cosmic radiation in planes (Darley, 2020). In other countries, a prospective research has used a commercially-available NN to fit the ground dose rate to the count rate measured by flying drones (most easily acquired) over Fukushima Daichi power plant (Sasaki, 2021) and radon dispersion based on environmental parameters has been predicted accurately (Duong, 2021).

Metamodelling is an innovation introduced by ML with applications in RP. A metamodel captures the physic of a phenomenon by learning from results obtained under various configurations and then predict (rather than calculate) a most foreseeable result, with less calculation and time, important in incidental conditions. Training set can come from experimental data or, more conveniently, simulation (CFD, Monte Carlo). Researches by French teams include the prediction of the dispersion of atmospheric pollutants/radionuclides at short range (Nony, 2021), in city streets (Mendil, 2021) and in the Fukushima region (Korsakissok, 2020). Metamodel has been used likewise for the dispersion of uranium in water (Lopez, 2021).

Globally, ML can find a fertile ground in all the complex/statistical phenomenon currently tackled by Monte Carlo calculation (Makovicka, 2009) and phenomenon with epistemic uncertainties (French, 2020). The fields of chemical and material science are getting a lot of traction for using ML model to predict the degradation of steels and to develop alloys for power plants [Mamun, 2021] and the nuclear domain might go this way.

### 4.3 Regulation

The official dose monitoring of workers is performed with passive dosimeters and active dosimeters are carried as a secondary device. Mentzel demonstrated that a NN trained on the glow curves of passive dosimeter can assess without the active dosimeter the date of a single high-dose irradiation; the method could be "*extended to a wide range of additional insights into irradiation scenarios*" (Mentzel, 2021).

In France, the SIANCE and PIREX projects, respectively led by the French nuclear safety authority (ASN) and the RP and Nuclear Safety Institute (IRSN), rely on Natural Language Processing (NLP) to account for the content of inspection follow-up reports written by ASN inspectors and of operating experience reports produced by operators after any safety or RP incidents (Boina, 2019, IRSN, Press 2021b). The two projects aim at constituting an intelligent database improving the analysis of data coming from inspections and operating experience. The NLP tool allows to have information more accessible, considering components, repetitive issues, trends and weak signals to define specific control tasks and confirm the guidance of inspection program (SIANCE) or direct expertise and R&D project (PIREX).

## 5. DISCUSSIONS

From the examples in the above part, it is noticeable that ML has been adapted to a wide-type of data generated from the RP disciplines and used in the three types of exposure situations under ICRP terminology (planned, existing and emergency) which epitomized how a highly flexible technology ML is. Plus, after the learning phase, ML uses a classical computer to provide outputs in very short time.

Globally, ML has benefited to RP in different ways:
- Facilitate and augment the expertise of human merely in classification tasks, free RP professionals' time to concentrate on problematic cases,
- Enable faster alert (anomaly, event) and assist in decision-making in providing fast results with reduced uncertainties,
- Introduce novel approach in capturing complex relationships and extract characteristics that may evade human capacities, notably in radiobiology, epidemiology and environmental modelling; where ML might be an interesting option (if not the only option) to grasp the complexity of the interactions.

The current status of most cases identified here is chiefly thesis, PhD and feasibility study, except for image interpretation in the medical field which is more industrialized. According to Figure 1, a more habitual usage of ML in RP could be expected, but not before several years. And a few issues need to be considered.

First, ML needs plenty of data, to such extend that "*the quality of the data is more important than the algorithm it-self*" according to Data Scientists. The preparation of the data relies on non-negligible human (expert) time, especially for the labelling of data for supervised learning (Boina, 2019). The control of data store needs to be considered both from an ethical and legal (ex. General Data Protection Regulation, GDPR, in Europe) points of view, noting that rules about data management in AI context are at the preface in Europe (EU-LEX) and that the community is also developing data privacy tool (ex. the open-source

OPACUS, released in December 2021). Data control might be a more pressing issue in the RP disciplines where medical, dosimetric, environmental etc. data are collected and cannot be widely shared.

Each algorithm is designed for a purpose and does not fall in the realm of the one pill solution. Therefore, the application of ML should start from the problem that need to be solved (and not the opposite). In most cases, the ML tool will be integrated in existing system and workflow, supporting the need to evaluate the implications on several dimensions such as benefits, limits, cost, time, practicability and interface (Orchard, 2020).

But while each tool is adapted to its objective, ensuring in the same time the reproducibility of the ML model is very important to validate the results, share them to others and avoid unnecessary duplication. In response to these challenges, it is possible to make the data FAIR (Findable, Accessible, Interoperable, Reusable) even in the context of ML (Wilkinson, 2016). In the biomedical sector, the developer's community has engaged the development of consistent (electronic) report on the characteristics of the ML techniques and performances in an explicit and transparent manner (ex. DOME and AIMe initiatives); making up the bases of standards that can be extended and improved with the upcoming advances of ML medical applications (Nature, 2021).

Indeed, the medical sector has been in forefront in ML and can be a compass for the other RP sectors. A fear expressed by medical imaging professionals was "*will a robot take my job?*" (Lewis, 2019) and although it seems it was not the case in practice, AI has raised a preoccupation about work rationalisation and standardisation (Mathieu-Fritz, 2021). Further, the usages of ML have generated questions with respect to the opacity of the algorithm and the explainability of the result, generating conflict where AI and human disagree (Anichini, 2021). The 'black-box syndrome' is not new (it probably came along with the usage of computer) but it is fundamental to re-interrogate it if the usage of AI in RP develops and especially if it is envisaged to engage AI in high-stake decision where human have prevailed until now, such as "*create complex treatment plan*" (EFOMP, 2021) or "*case-based reasoning tool*" in radiological emergency preparedness (Raskob, 2016). Furthermore, some areas of the RP job are based on implicit practices and good judgement (Lecomte, 2019), places where current AI is not very accurate. All these could give rise for consideration and principles where AI might be suitable (vs. not suitable) when tackling RP issues, based on the assessment of the implications prior to rollout.

As good practice, Data Scientists integrate confidence score and interpretability tools to build explainable AI (XAI [Tjoa, 2015]) for the user and guidelines in this field are under development by institution (HLEG, 2019). Nonetheless, some medical professional societies consider these are not sufficient and call for standards in the design of robust AI system, with validation, including ethical checks, by independent structure (RANZCR, 2019).

## 6. CONCLUSION

Artificial Intelligence, Machine Learning, Neural Network… a cloud of new words has emerged in the media and scientific literature and their semantic is varying from dream to nightmare. The objective of this article was first to go through the definition of Machine Learning (ML) to decipher what lies beneath the algorithms. Then, by using a bibliometric analysis, we showed that ML have been barely applied in RP yet, except in the medical sector.

Several publications on the applications of ML in RP have been analysed and the results presented with regard to the sector and the objectives. The achievements appear more mature in the medical field versus at the stage of R&D in others. Nonetheless ML has already shown their capability to get the same amount of expertise than human (but cheaper) thus use as a tool for repetitive tasks and detection; deal with a large amount of data with a little bit less of expertise than human but in a matter of second instead of days for assistance in decision-making by prediction; and build model with many more free parameters (variables) too complex for human mind. Their advantages rely on their remarkable flexibility and to adapt 'themselves' via a learning phase to the data.

The computing industry is among one of the fastest growing industry; as AI tool will become more common (and more affordable), it is foreseeable that AI technologies will become more usual in RP and more usages will develop. Some topics have been barely grappled, such as making use of dosimetric database or shielding procedures.

However, looking at the feedback from the pioneering medical sector, a number of key issues should be addressed before attempting to import ML heuristic in all RP: by exploring the evaluation of the advantages, considering data control, robustness and interpretability of the results and considering the ethical and legal implications of their uses.

For the effective application of ML in RP, we promote discussion and collaboration between the RP professionals and Data Scientists, in formal manner when developing AI tool and also in unformal forums. Cross-disciplines interactions will indeed support the acceptance of this technology, ensure that the developments are fit-for-purposes and the diffusion of the achievements. After realizing the real potential, this could accelerate innovation and research with benefits for radiation protection and maybe help to rejuvenate the sector.

## Acknowledgement

The authors wish to thank the Members of the Young Club of SFRP for collecting documents about AI, substantiating the work and critical review. Members of SFRP Administrative Board and Environment Section have also provided useful insights and contacts. The authors would like to specially thanks the RP Professionals and the Data Scientists at ASN, IRSN (ENV, LEPI) and Icohup for their time in (virtual) meeting and answering our questions by mail or phone.